\documentclass[letterpaper, 10 pt, conference]{ieeeconf}

\IEEEoverridecommandlockouts
\usepackage[english]{babel}
\usepackage{amsmath}
\usepackage{amssymb}
\usepackage{booktabs}
\usepackage[dvips]{graphicx}
\usepackage{multirow}
\usepackage{amsfonts}
\usepackage{enumerate}
\usepackage{tabularx}
\usepackage{algorithm,algorithmic}
\usepackage{bm}
\usepackage{enumerate}
\usepackage{adjustbox}
\usepackage{xcolor}
\usepackage{siunitx}
\usepackage{booktabs}
\usepackage{mdframed}
\usepackage{adjustbox}
\usepackage{authblk}
\usepackage{color}
\usepackage{xurl}
\usepackage{subcaption}
\usepackage{cite}
\makeatletter
\let\NAT@parse\undefined
\makeatother
\usepackage{hyperref}
\usepackage{relsize}
\usepackage{float}
\usepackage{pifont}

\let\labelindent\relax
\usepackage{enumitem}
\setlist[itemize]{label=\rule[0.5ex]{0.6ex}{0.6ex}}

\newcommand{\cmark}{\textcolor{green}{\ding{51}}}%
\newcommand{\xmark}{\textcolor{red}{\ding{55}}}%

\usepackage{stackengine,scalerel}

\usepackage{fontawesome5}

\title{\LARGE \bf Learning Long-Horizon Predictions for Quadrotor Dynamics}

\author{ Pratyaksh Prabhav Rao$^1$, Alessandro Saviolo$^1$, Tommaso Castiglione Ferrari$^2$, and Giuseppe Loianno$^1$

\thanks{$^1$The authors are with the New York University, Tandon School of Engineering, Brooklyn, NY 11201, USA. {\tt\footnotesize email: \{pr2257, as16054, loiannog\}@nyu.edu}.}
\thanks{$^2$The author is with the Autonomous Robotics Research Center-Technology Innovation Institute, Abu Dhabi, UAE. {\tt\footnotesize email:  \{Tommaso.Castiglione\}@tii.ae}.}
\thanks{This work was supported by the Technology Innovation Institute, the NSF CAREER Award 2145277, and the DARPA YFA Grant D22AP00156-00. Giuseppe Loianno serves as consultant for the Technology Innovation Institute. This arrangement has been reviewed and approved by the New York University in accordance with its policy on objectivity in research.}
}

\begin{document}


\maketitle


\begin{abstract}
Accurate modeling of system dynamics is crucial for achieving high-performance planning and control of robotic systems. Although existing data-driven approaches represent a promising approach for modeling dynamics, their accuracy is limited to a short prediction horizon, overlooking the impact of compounding prediction errors over longer prediction horizons. Strategies to mitigate these cumulative errors remain underexplored. To bridge this gap, in this paper, we study the key design choices for efficiently learning long-horizon prediction dynamics for quadrotors. Specifically, we analyze the impact of multiple architectures, historical  data, and multi-step loss formulation. We show that sequential modeling techniques showcase their advantage in minimizing compounding errors compared to other types of solutions. Furthermore, we propose a novel decoupled dynamics learning approach, which further simplifies the learning process while also enhancing the approach modularity.
Extensive experiments and ablation studies on real-world quadrotor data demonstrate the versatility and precision of the proposed approach. Our outcomes offer several insights and methodologies for enhancing long-term predictive accuracy of learned quadrotor dynamics for planning and control.  
\end{abstract}


\section*{Supplementary Material}
\noindent \textbf{Video}: \url{https://youtu.be/MPUJunMD11U}
\\
\noindent \textbf{Code}: \url{https://github.com/arplaboratory/long-horizon-dynamics}

\section{Introduction} \label{sec:introduction}
Unmanned Aerial Vehicles (UAVs), including quadrotors, are becoming integral to a variety of applications, including logistics, reconnaissance missions, search and rescue, and inspections scenarios~\cite{rao2022quadformer}.
These tasks require UAVs to precisely navigate through unknown cluttered environments, which demands planning collision-free paths and controlling the UAV to closely follow these paths~\cite{kulkarni2024reinforcement}. 
The effectiveness of both planning and control critically relies on the accurate prediction of action sequence outcomes, necessitating precise system dynamics modeling~\cite{saviolo2023learning}. 
Yet, modeling these dynamics is often challenging due to complex aerodynamic forces, interactions between propellers, and other nonlinear phenomena experienced during different operating conditions, which traditional physics-based models often fail to capture accurately~\cite{saviolo2022physics, bauersfeld2021neurobem}.
These limitations can result in suboptimal flight performance and, eventually, catastrophic failures.

\begin{figure}[t]
    \includegraphics[width=\linewidth, trim=0 270 600 0, clip]{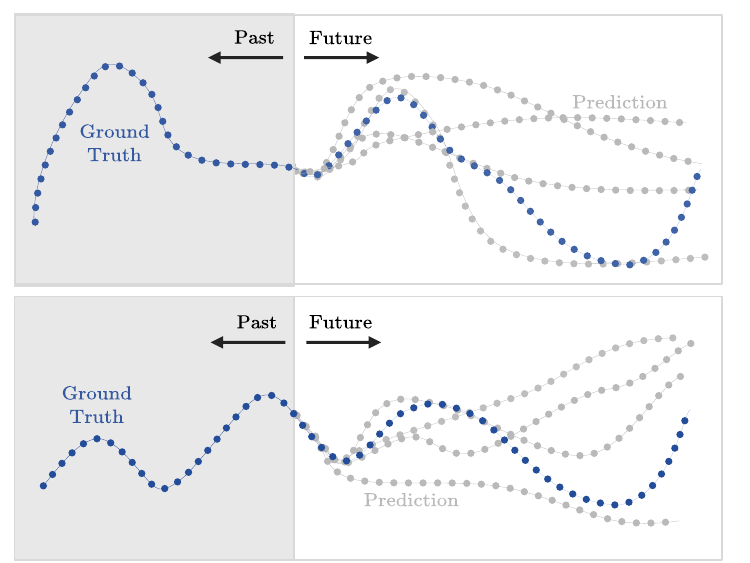}
    \caption{Illustration of how errors in dynamical models accumulate over time. Each subsequent prediction integrates the errors from all previous steps, leading to increased cumulative error and reduced accuracy in long-term predictions.}
    \label{fig:initial_figure}
      \vspace{-10pt}
\end{figure}

Recent advances have seen a shift towards data-driven approaches for modeling system dynamics, offering promising improvements in flight performance through offline learning and online adaptation~\cite{saviolo2023active, crocetti2023gapt, o2022neural}. Nonetheless, existing methodologies primarily focus on short-term predictive models, overlooking the significance of long-horizon predictions. Forecasting over long horizons  is necessary for effective planning and high performance control of robotic systems. For instance, long-horizon predictive capability enables planning algorithms to anticipate how the system will behave under long sequences of control actions, a crucial aspect for executing complex maneuvers precisely. Moreover, in the context of optimal control and model-based reinforcement learning~\cite{lambert2021learning}, accurate long-horizon predictions allow agents to foresee future actions that lead to maximization of expected cummulative rewards. Despite leveraging learned dynamics models for planning or control tasks by recursively applying them to forecast long-horizon trajectories, current methodologies often fall short due to the compounding error phenomenon, where each subsequent prediction incorporates all past errors, leading to cumulative inaccuracies over time (Figure~\ref{fig:initial_figure}). While prior studies have acknowledged the challenge of compounding errors~\cite{clavera2018model, wang2019benchmarking, heess2015learning, janner2019trust, venkatraman2015improving, asadi2019combating, xiao2019learning, lambert2021learning, lambert2022investigating}, comprehensive strategies to address it, particularly through design and architectural considerations for long-term predictive accuracy, remain largely unexplored. Our research aims to address this critical gap. We propose and and evaluate various model architectures tailored to sequential modeling tasks and design choices to enhance the expressiveness of learning-based dynamics for long-term prediction capability.



The key contributions can be summarize as follows. First, we propose and analyze several key design choices in the learning process, including historical data and multi-step loss formulation, to enhance long-term prediction accuracy. We demonstrate that employing sequential modeling techniques for quadrotor system dynamics is particularly advantageous. These techniques excel at minimizing compounding errors by accurately representing time-correlated features.
Second, we propose a novel decoupled dynamics learning framework that unlike conventional frameworks that directly forecast complete state information, breaks down system dynamics into manageable subproblems. This promotes  modularity and enables independent optimization for each component to enhance long-horizon prediction capability.
Finally, we conduct extensive experiments and ablation studies of the framework across diverse real-world quadrotor data. These experiments demonstrate the versatility and predictive precision of the framework in real-world scenarios.

This refined focus on sequential modeling, alongside a methodical examination of design choices, sets our work apart, offering new insights and methodologies for improving the performance of UAV planning and control problems.

\section{Related Works} \label{sec:related_works}
\subsection{Dynamics Learning for One-step Forecasts}
One-step dynamics learning models have proven to be highly effective in addressing a diverse array of robotics tasks. 
For instance, Gaussian Processes (GPs) have successfully tackled various lower-dimensional robotic learning challenges, demonstrating their proficiency in managing uncertainty in a structured manner~\cite{koller2018learning, torrente2021data}.
However, GPs face scalability limitations, especially with tasks involving high dimensions and large datasets. 
On the contrary, deep neural networks have exhibited remarkable scalability to higher dimensions and the ability to handle large amounts of data effectively.
For instance, \cite{punjani2015deep} adopted a Multi-Layer Perceptron (MLP) to capture helicopter dynamics.
\cite{bansal2016learning} employed a shallow MLP to learn the full system dynamics of a quadrotor. 
Moreover, a diverse array of architecture types, incorporating sequence modeling techniques, has found practical utility in the realm of learning robot dynamics. 
Examples include the application of Recurrent Neural Networks (RNNs)~\cite{mohajerin2019multistep} and Temporal Convolutional Networks (TCNs)~\cite{kaufmann2020deep, looper2022temporal, saviolo2022physics}. Furthermore, other data-driven methods such as structured mechanical models~\cite{gupta2020structured} and Lagrangian networks~\cite{cranmer2020lagrangian} leverage deep learning to satisfy smooth constraints.
Despite the versatility offered by all these models, a common challenge arises, particularly in tasks requiring long-horizon planning and control, where they often encounter compounding errors. 

\subsection{Compounding Errors in Multi-step Forecasts}
The compounding error problem has been previously studied under the context of model-based reinforcement learning~\cite{clavera2018model, wang2019benchmarking}, where a dynamic model of the system is iteratively learned and recursively applied to derive a control policy.
For instance, \cite{heess2015learning} addresses compounding error in model prediction using real observations, aiming to avoid distribution drift. 
\cite{janner2019trust} tackles the problem with short horizons, at the expense of long-term capabilities. 
Other approaches involve tweaking model optimization, including imitation-learning-inspired models~\cite{venkatraman2015improving}, multi-step estimators~\cite{asadi2019combating}, and flexible prediction horizons~\cite{xiao2019learning}.

The model proposed by~\cite{lambert2021learning} introduces a new training paradigm to mitigate compounding error by embedding time dependence in predictions. 
However, it is currently limited by its requirement for closed-form controllers. 
Recently, \cite{lambert2022investigating} investigated various factors that influence the magnitude of long-term prediction error. 
Yet, this work primarily aims at understanding the properties and conditions causing compounding errors. 
The challenge of compounding error remains not fully understood in terms of design choices, techniques, and model architectures for enhancing long-horizon predictions of learning dynamic models. 
To address this gap, we present a specific set of network design and training choices to mitigate this challenging problem.

\begin{figure*}[t]
    \centering
    \includegraphics[width=\textwidth, trim=75 170 75 150, clip]{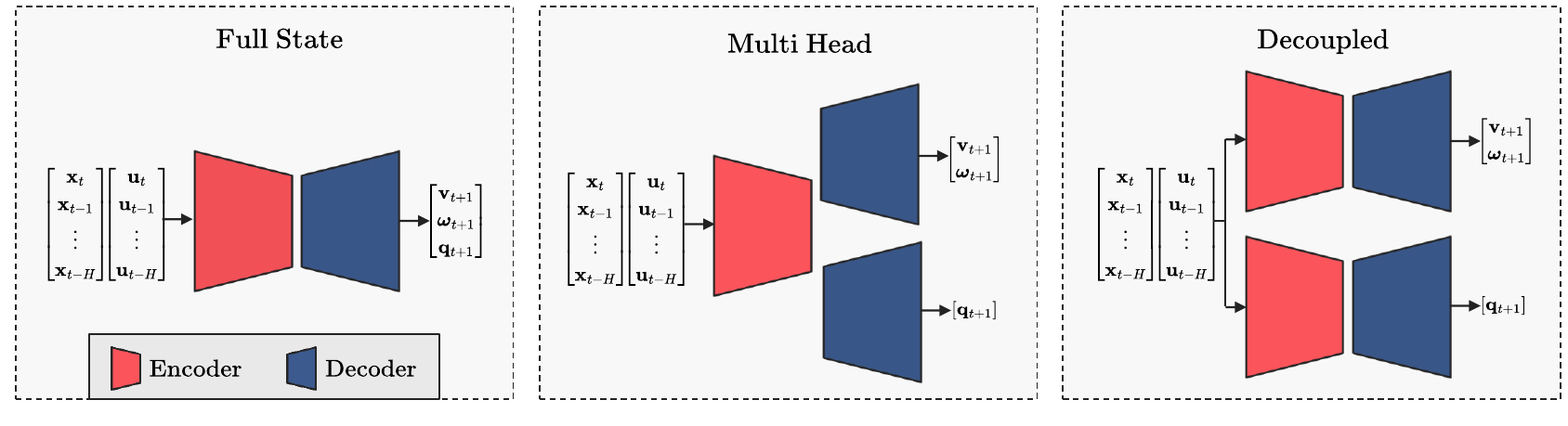}
    \caption{Full State Predictor predicts complete state vectors, Multi Head Predictor uses separate heads for velocity and attitude prediction, while our proposed Decoupled Predictor decomposes the problem into velocity and attitude prediction modules, introducing a novel approach for enhanced modularity and long-horizon prediction.}
    \label{fig:predictors}
    \vspace{-20pt}
\end{figure*}

\section{Background} \label{sec:background}
\subsection{Modeling the System Dynamics}
Consider the system's state $\mathbf{x}_{t} \in \mathbb{R}^{n}$ at time $t$, influenced by the action $\mathbf{u}_{t} \in \mathbb{R}^{m}$. 
Modeling the system dynamics requires finding a function $f : \mathbb{R}^{n} \times \mathbb{R}^{m} \rightarrow \mathbb{R}^{n}$ such that
\begin{equation}
    \mathbf{x}_{t+1} = f(\mathbf{x}_{t}, \mathbf{u}_{t})
    .
\end{equation}
The quadrotor's state at a given time index is given by $\arraycolsep=2pt \mathbf{x}_t=\begin{bmatrix}\mathbf{p}^\top_{t}&\mathbf{v}^\top_{t}&\mathbf{q}^\top_{t}&\boldsymbol{\omega}^\top_{t}\end{bmatrix}^\top$, where $\mathbf{p}_{t} \in \mathbb{R}^{3}$ and $\mathbf{v}_{t} \in \mathbb{R}^{3}$ are the robot's position and velocity expressed in the inertial frame, $\mathbf{q}_{t} \in \mathbb{R}^{4}$ is the robot's attitude using the unit quaternion representation with respect to the inertial frame, and $\boldsymbol{\omega}_{t} \in \mathbb{R}^{3}$ is the robot's angular velocity in the body frame. 
Furthermore, the control action is represented by $\mathbf{u}_{t} \in \mathbb{R}^{4}$ and  corresponds to the motor speeds.

\subsection{Learning One-step System Dynamics}
The common practice for learning the system dynamics involves training a one-step predictive model $h_{\boldsymbol{\theta}}$, parametrized by $\boldsymbol{\theta}$, on a dataset of $N$ collected state-action trajectories $D = \{(\mathbf{x}_{i}, \mathbf{u}_{i}, \mathbf{x}_{i+1})\}_{i=1}^{N}$. The input to the neural network consists of linear velocity $\mathbf{v}$, angular velocity $\boldsymbol{\omega}$, attitude $\mathbf{q}$, and control inputs $\mathbf{u}$. Position information is omitted, assuming position-independent system dynamics, as the robot's positional changes can be recovered via Euler integration. The training process optimizes $\boldsymbol{\theta}$ to minimize the prediction error of $h_{\boldsymbol{\theta}}$ over $D$ as follows
\begin{equation}
    \min_{\boldsymbol{\theta}} 
    \frac{1}{N} 
    \sum_{D} 
    || \mathbf{x}_{t+1} - \hat{\mathbf{x}}_{t+1}||^2_2
    ,
\label{one-step}
\end{equation}
where $\hat{\mathbf{x}}_{t+1} = h_{\boldsymbol{\theta}}(\mathbf{x}_{t}, \mathbf{u}_{t})$. Rather than solely predicting the true state observation as $\hat{\mathbf{x}}_{t+1} = h_{\boldsymbol{\theta}}(\mathbf{x}_{t}, \mathbf{u}_{t})$, an alternative approach involves predicting the change in the current state, expressed as $\hat{\mathbf{x}}_{t+1} = \mathbf{x}_{t} + h_{\boldsymbol{\theta}}(\mathbf{x}_{t}, \mathbf{u}_{t})$. This technique, widely adopted~\cite{janner2019trust, yu2020mopo}, is popular in regularizing the prediction distribution.
Therefore, we employ it in this work.

\subsection{Compounding Errors in Multi-step Forecasts}
When forecasting the outcome for a given sequence of control actions $T$ steps in the future, the one-step dynamics model is recursively applied as
\begin{equation}
    \hat{\mathbf{x}}_{t+T} =
    h_{\boldsymbol{\theta}}
    ( \dots
        h_{\boldsymbol{\theta}}
        (
            h_{\boldsymbol{\theta}}(\mathbf{x}_t, \mathbf{u}_t),
        \mathbf{u}_{t+1}
        ) 
    \dots , \mathbf{u}_{t+T}
    )
    .
\end{equation}
However, any prediction error caused by inaccurately modeling the system dynamics, namely $\epsilon_{t} = ||\mathbf{x}_{t} - \hat{\mathbf{x}}_{t}||^2_2$, undergoes multiplicative growth due to each subsequent prediction's input being influenced by past errors. 
Formally, the compounding error problem in multi-step forecasts over a $T$ time horizon can be formulated as
\begin{equation}
    \begin{aligned}
        \hat{\mathbf{x}}_{t+1} &= 
        h_{\boldsymbol{\theta}} (\mathbf{x}_t, \mathbf{u}_t) 
        + \epsilon_t,\\
        \hat{\mathbf{x}}_{t+2} &= 
        h_{\boldsymbol{\theta}} (\hat{\mathbf{x}}_{t+1}, \mathbf{u}_{t+1}) 
        + \epsilon_{t+1},\\
        &\vdots\\
        \hat{\mathbf{x}}_{t+T} &= 
        h_{\boldsymbol{\theta}} (\hat{\mathbf{x}}_{t+T-1}, \mathbf{u}_{t+T-1}) 
        + \epsilon_{t+T}.
    \end{aligned}
\end{equation}
This compounding effect of errors is a consequence of each subsequent prediction incorporating all past errors, leading to a cumulative effect on the overall prediction accuracy.

\section{Methodology} \label{sec:methodology}
\subsection{Model Architectures}
The inherent challenge in data-driven dynamics learning lies in the degradation of state and action information by sensor noise. Consequently, the Markovian assumptions on the robot dynamics and full observability are constrained. Recognizing this, previous studies explored the integration of historical information to address these limitations. Leveraging recently observed states and actions, which retain redundant patterns, provides a mechanism for data-driven models to mitigate the effects of noise~\cite{saviolo2022physics}. Formally, this integration involves histories of states $\mathbf{X}_{t} = [\mathbf{x}_{t-H}^{\top} \dots \mathbf{x}_{t}^{\top}]^{\top}$, and control inputs, denoted as $\mathbf{U}_{t} = [\mathbf{u}_{t-H}^{\top} \dots \mathbf{u}_{t}^{\top}]^{\top}$, both with a length of $H$, enabling the prediction of the state at time $t+1$ as $\hat{\mathbf{x}}_{t+1} = \mathbf{x}_{t} + h_{\boldsymbol{\theta}}(\mathbf{X}_{t}, \mathbf{U}_{t})$. 

While historical information has demonstrated effectiveness in learning accurate dynamics, its potential with different model architectures tailored to capturing long-range time dependencies remains largely unexplored, particularly in addressing the compounding error problem. Traditional MLPs, commonly employed for such tasks, struggle to leverage temporal context effectively due to inherent architectural limitations. This leads to inaccuracies and high variance in predictive capability. To bridge this gap, our study focuses on benchmarking several state-of-the-art recurrent architectures, including Long Short-Term Memory (LSTM) \cite{hochreiter1997long}, Gated Recurrent Unit (GRU) \cite{chung2014empirical}, and TCN \cite{lea2017temporal}. LSTM and GRU, being variants of RNNs, are specifically designed to capture long-range dependencies in sequential data. Specifically, LSTMs incorporate memory cells and multiple gating mechanisms, while GRUs simplify this architecture by combining gates. While LSTMs are computationally expensive, GRUs offer performance with lower computational complexity. TCNs, leveraging causal convolutions, provide efficient training and scalability. However, they may require more data for optimal performance.

\subsection{Multi-Step Loss}
Recent works on learning dynamics models~\cite{saviolo2022physics, bauersfeld2021neurobem} utilize a single step loss, where the model is trained to predict the immediate next state. The loss function is computed as shown in eq.~(\ref{one-step}). This formulation focuses on short-term prediction accuracy and often fails in applications involving long-horizon planning and control. To tackle this problem, recent approaches~\cite{clavera2018model, wang2019benchmarking, lambert2021learning} have adopted a multi-step loss formulation which improves the long-term predictive capability. The multi-step loss formulation involves predicting multiple future states beyond just the immediate next step. The model is trained to forecast the system's behavior over a longer horizon by recursively predicting $U$ future steps.
The loss function is computed based on the cumulative error over all predicted future steps compared to their corresponding actual future states as
\begin{equation} \label{eq:multi_step_loss}
    \min_{\boldsymbol{\theta}} 
    \frac{1}{U N} 
    \sum_{D} 
    \sum_{i=1}^{U} 
    \Vert\mathbf{x}_{t+i} - \hat{\mathbf{x}}_{t+i}\Vert^2_2
    .
\end{equation}
This approach provides a more comprehensive evaluation of the model predictive performance over longer time horizons and is beneficial for tasks requiring foresight and planning.

\begin{table*}[t]
    \setlength{\tabcolsep}{4.9pt}
    \centering
    \caption{Impact of historical data and multi-step loss on long-term prediction. TCN achieves superior performance with a history of $20$ and a multi-step horizon of $10$ for loss computation. All models use the decoupled predictor type and are evaluated over $60$ time steps on unseen trajectories, averaging results over $3$ training runs with different seeds. Blank entries denote that LSTM, GRU, and TCN, tailored for sequential data, exclude history information, indicating no sequence data.}
    \vspace{-0.25em}
    \begin{tabular}{cccccccccccccccccc}
        \toprule\toprule
        \multirow{3}{*}{H} & \multirow{3}{*}{U}
        & \multicolumn{8}{c}{PI-TCN \cite{saviolo2022physics}} & \multicolumn{8}{c}{NeuroBEM \cite{bauersfeld2021neurobem}} \\
        \cmidrule(lr){3-10}\cmidrule(lr){11-18} &&
        \multicolumn{2}{c}{MLP} & \multicolumn{2}{c}{LSTM} & \multicolumn{2}{c}{GRU} & \multicolumn{2}{c}{TCN}  &
        \multicolumn{2}{c}{MLP} & \multicolumn{2}{c}{LSTM} & \multicolumn{2}{c}{GRU} & \multicolumn{2}{c}{TCN}  \\
        \midrule
         && $\delta_{\mathbf{v}}$ & $\delta_{\mathbf{q}}$ & $\delta_{\mathbf{v}}$ & $\delta_{\mathbf{q}}$ 
        & $\delta_{\mathbf{v}}$ & $\delta_{\mathbf{q}}$ & $\delta_{\mathbf{v}}$ & $\delta_{\mathbf{q}}$
        & $\delta_{\mathbf{v}}$ & $\delta_{\mathbf{q}}$ & $\delta_{\mathbf{v}}$ & $\delta_{\mathbf{q}}$ 
        & $\delta_{\mathbf{v}}$ & $\delta_{\mathbf{q}}$ & $\delta_{\mathbf{v}}$ & $\delta_{\mathbf{q}}$ \\
        \midrule
        $1$ & $1$ &
        $1.325$ & $0.366$ & $-$ & $-$ & $-$ &
        $-$ & $-$ & $-$ &  
        $1.335$ & $0.334$ & $-$ & $-$ & $-$ 
        & $-$ & $-$ & $-$  \\
        $5$ & $1$ &
        $0.872$ & $0.301$ & $0.326$ & $0.289$ & $0.467$ &
        $0.278$ & $0.192$ & $0.140$  & 
        $0.808$ & $0.103$ & $0.489$ & $0.089$ & $0.468$
        & $0.083$ & $0.201$ & $0.072$  \\
        $5$ & $5$ &
        $0.722$ & $0.289$ & $0.318$ & $0.240$ & $0.426$ &
        $0.243$ & $0.184$ & $0.112$ & 
        $0.728$ & $0.098$ & $0.402$ & $0.081$ & $0.442$ 
        & $0.078$ & $0.199$ & $0.066$  \\
        $5$ & $10$ &
        $0.532$ & $0.277$ & $0.207$ & $0.202$ & $0.365$ &
        $0.200$ & $0.180$ & $0.102$  & 
        $0.688$ & $0.090$ & $0.389$ & $0.079$ & $0.399$
        & $0.076$ & $0.188$ & $0.060$ \\
        $10$ & $1$ &
        $0.780$ & $0.200$ & $0.210$ & $0.143$ & $0.415$ &
        $0.156$ & $0.156$ & $0.087$  & 
        $0.487$ & $0.077$ & $0.281$ & $0.063$ & $0.300$ 
        & $0.665$ & $0.128$ & $0.046$ \\
        $10$ & $5$ &
        $0.692$ & $0.160$ & $0.204$ & $0.114$ & $0.388$ &
        $0.100$ & $0.154$ & $0.064$  & 
        $0.365$ & $0.056$ & $0.221$ & $0.054$ & $0.278$ 
        & $0.059$ & $0.112$ & $0.036$  \\
        $10$ & $10$ &
        $0.546$ & $0.121$ & $0.200$ & $0.091$ & $0.316$ &
        $0.098$ & $0.132$ & $0.054$  & 
        $0.307$ & $0.049$ & $0.145$ & $0.050$ & $0.207$ 
        & $0.047$ & $0.094$ & $0.025$  \\
        $20$ & $1$ &
        $0.487$ & $0.100$ & $0.155$ & $0.066$ & $0.302$ &
        $0.082$ & $0.102$ & $0.034$  & 
        $0.234$ & $0.039$ & $0.101$ & $0.030$ & $0.172$
        & $0.036$ & $0.078$ & $0.012$  \\
        $20$ & $5$ &
        $0.499$ & $0.098$ & $0.124$ & $0.055$ & $0.263$ &
        $0.073$ & $0.090$ & $0.023$ & 
        $0.183$ & $0.030$ & $0.091$ & $0.027$ & $0.125$ 
        & $0.025$ & $0.055$ & $0.007$  \\
        $20$ & $10$ &
        $0.357$ & $0.088$ & $0.089$ & $0.048$ & $0.138$ &
        $0.062$ & $\textbf{0.062}$ & $\textbf{0.016}$ & 
        $0.125$ & $0.025$ & $0.077$ & $0.011$ & $0.090$ 
        & 0.010 & \textbf{0.042} & \textbf{0.006}\\
        \bottomrule\bottomrule
    \end{tabular}
    \label{tab:ablation}
    \vspace{-10pt}
\end{table*}

\subsection{Dynamics Decoupling}
The formulation of learning dynamics models involves various strategies, among which the full-state predictor and the multi-head predictor are prominent (Figure~\ref{fig:predictors}). The full-state predictor aims to directly forecast the complete state vector, encompassing linear velocity $\mathbf{v}$, angular velocity $\boldsymbol{\omega}$, and attitude $\mathbf{q}$, offering a comprehensive view of system dynamics. However, its holistic approach may encounter challenges in capturing long-term dependencies due to the complexity of underlying dynamics and the high-dimensional output space, potentially increasing sensitivity to data noise.

Conversely, the multi-head predictor divides the prediction task into separate heads, typically focusing on velocity and attitude independently. While this specialization allows tailored modeling of different state aspects, it introduces coordination challenges. Dividing the prediction task can hinder effective prediction coordination, leading to inconsistencies and limited information sharing between the decoders, reducing predictive accuracy. In contrast, our proposed modular approach decouples system dynamics into manageable subproblems. By focusing on distinct components, such as velocity and attitude, decoupled predictors promote modularity, simplify learning and enable independent optimization. This strategy enhances the model's capability to capture complex dynamics and facilitates more accurate long-term predictions, addressing the limitations of traditional predictors. Specifically, we introduce two key modules: the Velocity Predictor and the Attitude Predictor. The Velocity Predictor is designed to forecast the change in velocity at the next time step. Formally, it is expressed as
\begin{equation}
    \hat{\mathbf{z}}_{t+1} = \mathbf{z}_{t} + h_{\boldsymbol{\theta}^{vel}}(\mathbf{X}_{t}, \mathbf{U}_{t}),
\end{equation}
where $\hat{\mathbf{z}}_{t+1}$ and $\mathbf{z}_{t}$ denote the predicted velocity and current velocity, encompassing both linear and angular changes - $\mathbf{z} = \begin{bmatrix} \mathbf{v} & \boldsymbol{\omega} \end{bmatrix}^{\top}$. On the other hand, the Attitude Predictor forecasts the change in attitude quaternion at the next time step and is formulated as
\begin{equation}
    \hat{\mathbf{q}}_{t+1} = h_{\boldsymbol{\theta}^{att}}(\mathbf{X}_{t}, \mathbf{U}_{t}) \odot \mathbf{q}_{t},
\end{equation}
where $\odot$ represents the quaternion-vector product, and $\hat{\mathbf{q}}_{t+1}$ and $\mathbf{q}_{t}$ denote the predicted and current attitude, respectively. 

\section{Experimental Setup} \label{sec:exp_setup}
\subsection{Datasets}
We extensively perform experiments on two well-known open-source real-world quadrotor datasets to analyze the long-term predictive performances of the neural models. 
\begin{figure}[t]
  \begin{center}
    \includegraphics[height=13.0em, trim=20 20 15 20, clip]{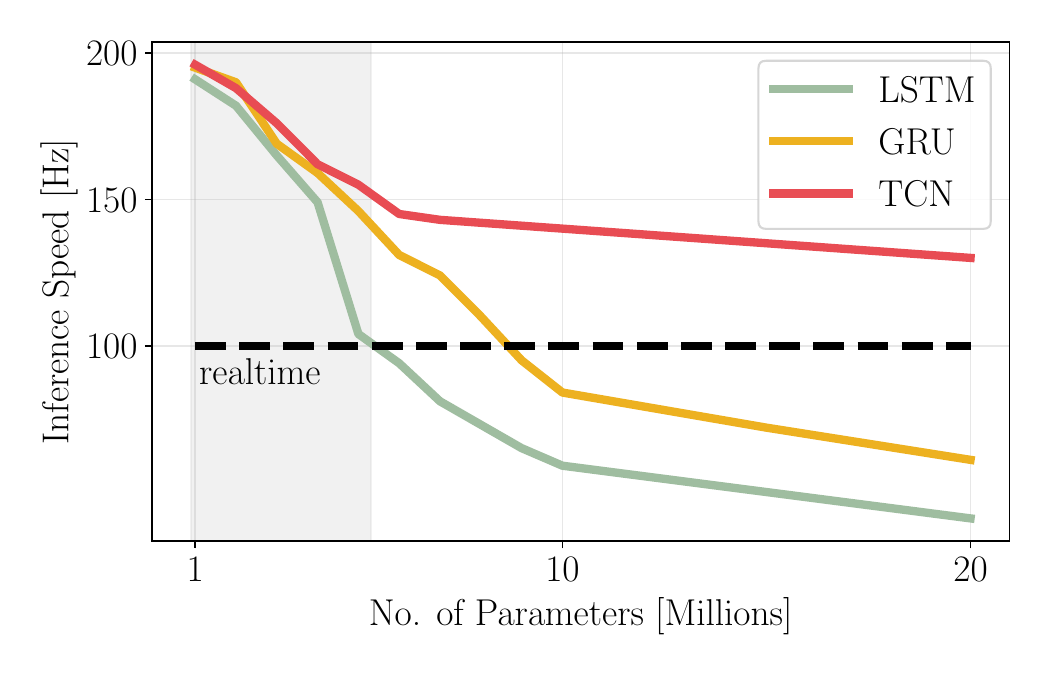}
    \caption{Selecting model parameters ensuring real-time performance on embedded systems. The parameter bound is determined by selecting the model with the lowest parameter count capable of real-time performance. LSTM achieves real-time predictions with up to $5.2$ million parameters. We freeze this bound across all models to have a fair comparison.}
    \label{inference_speed}
  \end{center}
  \vspace{-20pt}
\end{figure}

\textbf{PI-TCN}. 
This dataset~\cite{saviolo2022physics} includes $68$ trajectories with a total flight time of $58$ min $3$ sec. These cover a diverse range of motions, including straight-line accelerations, circular movements, parabolic maneuvers, and lemniscate trajectories. The dataset is designed to capture complex effects, pushing the quadrotor to its physical limits with speeds of 6 $\text{m}\text{s}^{-1}$, linear accelerations of $18~\text{m}\text{s}^{-2}$, angular accelerations of $54~\text{rad}\text{s}^{-2}$, and motor speeds of 16628 rpm. Data is sampled at $100$ Hz. We use $54$ trajectories for training, $10$ for validation, and $4$ for testing, ensuring a comprehensive evaluation across various challenging scenarios.

\textbf{NeuroBEM}. 
This dataset~\cite{bauersfeld2021neurobem} comprises $96$ flights with a total flight time of $1$ hr $15$ min, encapsulating the entire performance envelope of the platform up to observed speeds of $18$ $\text{m}\text{s}^{-1}$ and accelerations of $46.8$ $\text{m}\text{s}^{-2}$. While the original dataset is sampled at $400$ Hz, we resampled it at $100$ Hz for all experiments. We utilize $67$ trajectories for training, $17$ for validation, and $12$ for testing.

\begin{figure*}[t]
    \centering
    \begin{subfigure}[t]{0.5\textwidth}
    \centering
    \includegraphics[width=\linewidth, trim=0 0 0 0, clip]{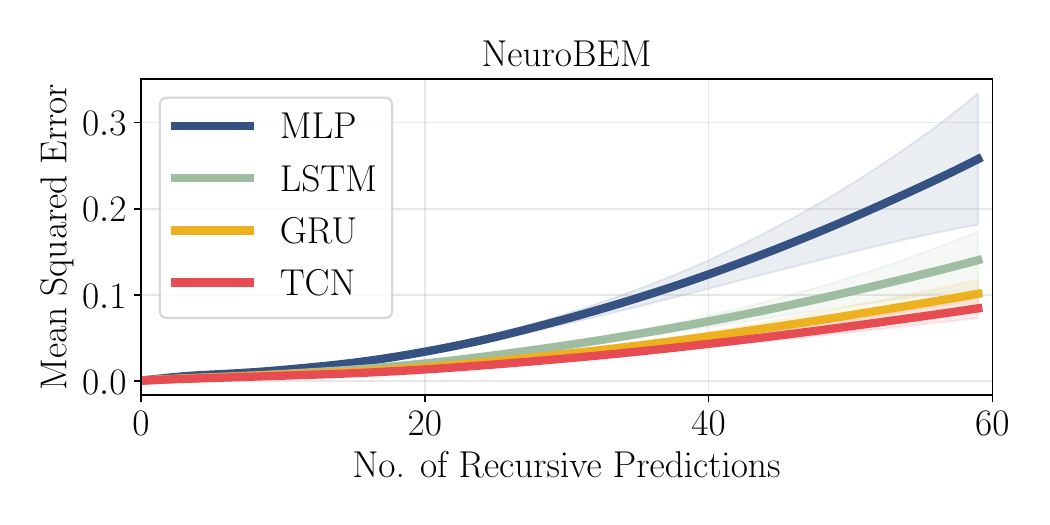}
    \end{subfigure}\hfill
    \begin{subfigure}[t]{0.5\textwidth}
    \centering
    \includegraphics[width=\linewidth, trim=0 0 0 0, clip]{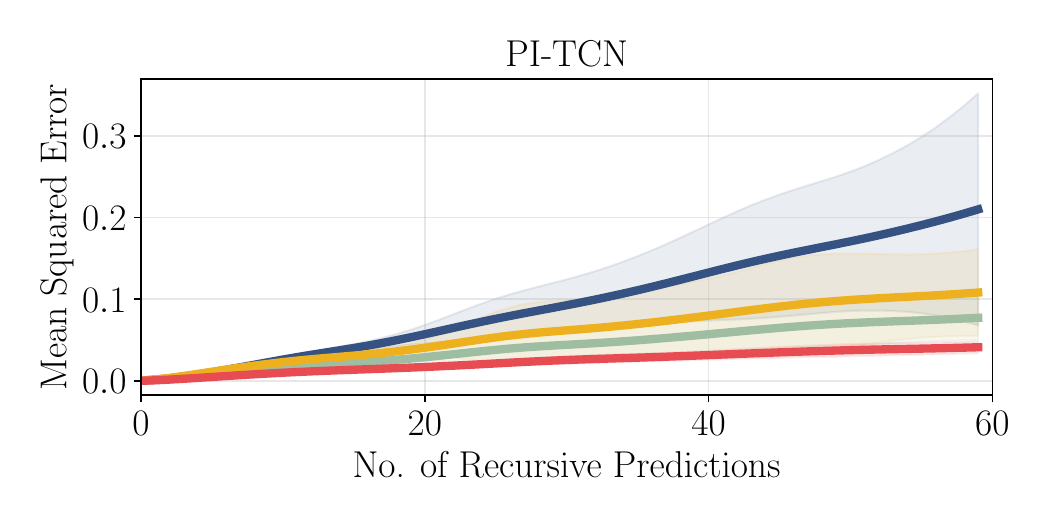}
    \end{subfigure}\hfill
    \vspace{-10pt}
    \caption{Mean Squared Error per-step among baseline neural networks models on unseen test trajectories. The MLP model demonstrates higher error variance across diverse unseen trajectories compared to other sequential models, attributed to the absence of architectural priors for capturing spatio-temporal dependencies across datasets.}
    \label{fig:analysis_recursion}
    \vspace{-10pt}
\end{figure*}

\subsection{Training}
We carefully chose the model architecture parameters to ensure real-time performance on an embedded platform. By analyzing the inference speed of baseline models relative to their parameter count (see Figure~\ref{inference_speed}), we establish a parameter bound by identifying the model with the lowest parameter count capable of real-time performance. LSTM demonstrates real-time predictions with up to $5.2$ million parameters. To ensure fair comparisons, we select this parameter bound across all models. All models adopt an encoder-decoder structure. For consistency, all encoders feature three layers. The MLP encoder consists of layers with $1024$, $512$, and $512$ neurons, respectively. Similarly, the LSTM and GRU encoders consist of three layers with $512$ neurons each in the hidden state. The TCN encoder integrates three hidden layers with sizes of $512$, $256$, and $256$ neurons, leveraging temporal convolutional layers with a LeakyReLU activation function, batch normalization, kernel size of $3$, and a dilation factor of $2$. All architectures incorporate an MLP decoder composed of three layers with $512$, $256$, and $256$ neurons. In our training process, we chose not to normalize the input state and actions (motor speed), as we notice no significant performance improvement. However, we scale the motor speed data by multiplying them by $10^{-3}$ to ensure equal distribution of data component scales, allowing the neural network to assign equal importance to all components. We employ the AdamW optimizer for training over $50$~K iterations, $\beta_1$ and $\beta_2$ set to $0.9$ and $0.999$, respectively, and a weight decay of $10^{-4}$. We train models with a batch size of $512$, constant learning rate warm up, lasting for $5$K iterations, followed by a cosine annealing learning rate scheduler. 

\subsection{Evaluation Metric}
We employ a sliding window approach with size $H$ along the unseen testing trajectory. At each state-control slice of $H$, the velocity predictor forecasts linear and angular velocities $T$ steps ahead, and we compute the velocity error between predicted and ground truth velocity values
\begin{equation}
\delta_{\mathbf{z}} = \frac{1}{T} \sum_{i=0}^{T-1} (\mathbf{z}_{i} - \hat{\mathbf{z}}_{i})^\top (\mathbf{z}_{i} - \hat{\mathbf{z}}_{i}),
\end{equation}
where \( \mathbf{z}_{i} \) represents the ground truth velocities at time index \( i\), and \( \hat{\mathbf{z}}_{i} \) denotes the predicted velocities at time index \( i \). Similarly, the attitude predictor forecasts unit quaternion states \( T \) steps ahead, and we compute the quaternion error with respect to the ground truth unit quaternion. We consider that the orientation is not an element of the Euclidean space~\cite{huynh2009metrics}. Therefore, to compute the quaternion error we take the logarithm of the rotation difference between the predicted and ground truth quaternion. The error is
\begin{equation}
\delta_{\mathbf{q}} = \frac{1}{T} \sum_{i=0}^{T-1} \theta_{i},
\end{equation}
where, for a given time index $i$, $\theta_{i}$ is calculated as
\begin{equation}
\theta_{i} = \arctan \left( \frac{\Vert\Delta \mathbf{q}^{\text{error}}_{i}\Vert}{\Delta \text{q}^{\text{error}}_{i}} \right).
\end{equation}
The terms $\Delta \mathbf{q}^{\text{error}}_{i}$ and $\Delta \text{q}^{\text{error}}_{i}$ denote the vector and scalar components of the quaternion respectively, and $\Delta \mathbf{q}^{\text{error}}_{i}$ is
\begin{equation}
\Delta \mathbf{q}^{\text{error}}_{i} = \mathbf{q}^{gt}_{i} \odot (\mathbf{q}^{pred}_{i})^{-1},
\end{equation}
where $\mathbf{q}^{gt}_{i}$ and $\mathbf{q}^{pred}_{i}$ represent the ground truth and predicted quaternions, respectively. Finally, $\log \Delta \mathbf{q}^{\text{error}}_{i} = \mathbf{u} \theta_{i}$, where $\mathbf{u} = \Delta \mathbf{q}^{\text{error}}_{i} / \Vert\Delta \mathbf{q}^{\text{error}}_{i}\Vert$. Both $\delta_{\mathbf{z}}$ and $\delta_{\mathbf{q}}$ are averaged across all slices of different testing trajectories to evaluate the model's predictive accuracy. We evaluate the ability to predict horizons of $60$ steps across the unseen testing trajectories of both datasets for all experiments. The reported experimental results are obtained by averaging the models trained with $3$ different random seeds, ensuring robustness and reliability, unless explicitly stated otherwise. 

\begin{table*}[!t]
    \setlength{\tabcolsep}{11.5pt}
    \centering
    \caption{Comparative performance of predictor types across trajectories with varied aggressiveness. The predictors utilize TCN architecture, a history length of $20$, and a multi-step horizon of $10$ for loss computation. Evaluation are conducted over $60$ time steps on unseen trajectories by averaging the results over $10$ training runs with different seeds.}
    \vspace{-0.25em}
    \begin{tabular}{clcccccccc}
    \toprule\toprule
    & \multirow{2}{*}{Trajectory} 
    & \multirow{2}{*}{$\mathbf{v}_{\text{mean}}$ [$\text{m}\text{s}^{-1}$]} 
    & \multirow{2}{*}{$\mathbf{v}_{\text{max}}$ [$\text{m}\text{s}^{-1}$]} 
    & \multicolumn{2}{c}{Full State} 
    & \multicolumn{2}{c}{Multi Head}
    & \multicolumn{2}{c}{Decoupled} \\
    \cmidrule(lr){5-6}\cmidrule(lr){7-8}\cmidrule(lr){9-10}
    &&&& $\delta_{\mathbf{v}}$ & $\delta_{\mathbf{q}}$ & $\delta_{\mathbf{v}}$ & $\delta_{\mathbf{q}}$ & $\delta_{\mathbf{v}}$ & $\delta_{\mathbf{q}}$  \\
    \midrule
    \multirow{8}{*}{\rotatebox[origin=c]{90}{PI-TCN \cite{saviolo2022physics}}}  & 
    Ellipse\_1 & $1.21$ & $1.59$ & $0.023$ & $0.034$ & $0.011$ & $0.019$ & $\textbf{0.007}$ & $\textbf{0.012}$\\
    & WarpedEllipse\_1 & $1.61$ & $2.44$ & $0.082$ & $0.083$ & $0.016$ & $0.021$ & $\textbf{0.008}$ & $\textbf{0.014}$ \\
    & Parabola & $2.36$ & $3.97$ & $0.112$ & $0.083$ & $0.102$ & $0.051$ &  $\textbf{0.039}$ & $\textbf{0.026}$ \\
    & ExtendedLemniscate & $2.44$ & $4.25$ & $0.321$ & $0.104$ & $0.564$ & $0.059$ & $\textbf{0.122}$ & $\textbf{0.032}$ \\
    & Lemniscate & $2.46$ & $4.73$ & $0.690$ & $0.089$ & $0.245$ & $0.076$ & $\textbf{0.050}$ & $\textbf{0.036}$  \\
    & WarpedEllipse\_2 & $2.90$ & $4.88$ & $0.135$ & $0.032$ & $0.107$ & $0.035$ & $\textbf{0.019}$ & $\textbf{0.010}$  \\
    & TransposedParabola & $2.73$ & $4.93$ & $0.431$ & $0.065$ & $0.650$ & $0.069$ & $\textbf{0.187}$ & $\textbf{0.019}$ \\
    & Ellipse\_2 & $3.27$ & $5.57$ & $0.266$ & $0.034$ & $0.192$ & $0.017$ & $\textbf{0.030}$ & $\textbf{0.015}$  \\
    \midrule
    \multirow{12}{*}{\rotatebox[origin=c]{90}{NeuroBEM \cite{bauersfeld2021neurobem}}} 
    & Lemniscate & $1.67$ & $3.51$ & $0.047$ & $0.042$ & $0.102$ & $0.129$ & $\textbf{0.017}$ & $\textbf{0.004}$  \\
    & Random Points & 2.38  & 8.25 & 0.182 & 0.089 & 0.211 & 0.101 & \textbf{0.076} & \textbf{0.007} \\
    & Lemniscate & $3.21$ & $7.04$ & $0.199$ & $0.031$ & $0.482$ & $0.015$ & $\textbf{0.094}$ & $\textbf{0.007}$  \\
    & Melon & $3.57$ & $7.63$ & $0.543$ & $0.034$ & $0.941$ & $0.028$ & $\textbf{0.107}$ & $\textbf{0.004}$  \\
    & Slanted Circle & $6.92$ & $10.75$ & $0.524$ & $0.029$ & $1.226$ & $0.116$ & $\textbf{0.140}$ & $\textbf{0.004}$  \\
    & Linear Oscillation & $7.25$ &  $16.95$ & $1.545$ & $0.043$ & $1.953$ & $0.078$ & $\textbf{0.214}$ & $\textbf{0.008}$  \\
    & Race Track & $7.64$ & $13.14$ & $2.993$ & $0.092$ & $3.656$ & $0.071$ & $\textbf{0.697}$ & $\textbf{0.006}$ \\
    & Melon & $7.74$ & $13.55$ & $1.842$ & $0.101$ & $2.921$ & $0.077$ & $\textbf{0.091}$ & $\textbf{0.004}$  \\
    & Slanted Circle & $8.57$ & $13.32$ & $0.598$ & $0.043$ & $1.206$ & $0.087$ & $\textbf{0.141}$ & $\textbf{0.002}$\\
    & Race Track & $9.94$ & $17.81$ & $4.434$ & $0.087$ & $5.024$ & $0.107$ & $\textbf{0.709}$ & $\textbf{0.109}$ \\
    & Lemniscate & $12.01$ & $19.83$ & $2.563$ & $0.054$ & $3.344$ & $0.106$ & $\textbf{0.711}$ & $\textbf{0.005}$ \\
    & Ellipse & $15.02$ & $19.20$ & $6.235$ & $0.065$ & $6.813$ & $0.073$ & $\textbf{1.528}$ & $\textbf{0.024}$  \\
    \bottomrule\bottomrule
    \end{tabular}
    \vspace{-20pt}
    \label{tab:agressive_flight_analysis}
\end{table*}

\section{Results}
\subsection{Impact of History Length and Multi-step Loss}
In this section, we investigate the impact of incorporating historical information and utilizing multi-step loss formulation on model performance, with all models in this experiment employing the decoupled predictor type. Subsequently, we conduct an ablation study to showcase the superior performance of the decoupled predictor type compared to the previously mentioned predictor types, thereby further validating our proposed approach. We begin by comparing MLP models with and without history information, highlighting the importance of temporal context in predictive modeling. By leveraging historical data, the model improves its ability to capture system dynamics, as evidenced by quantitative results (Table~\ref{tab:ablation}). Additionally, we investigate multi-step loss formulation's effectiveness in enhancing long-term predictive accuracy, presenting comparative analysis between single-step and multi-step loss functions. Our study reveals that optimizing error over multiple future time steps $U$ reduces prediction errors and improves model robustness for mitigating compounding errors over long-horizon predictions. Unroll length of $1$ serves as the baseline, representing single-step loss with no unrolling. Our findings suggest an optimal configuration of a history length of $20$ paired with an unroll length of $10$, and is selected for all subsequent experiments. Exceeding an unroll length of $10$ leads to training instabilities due to large gradient values. We also notice that if we go beyond a history length of $20$, the error increases. Beyond a certain history length, the relevance of past observations may diminish, and including excessively distant past information may introduce noise or irrelevant patterns, hindering the model's ability to generalize effectively to unseen data. Furthermore, we observe MLP's limitations in effectively extracting temporal context from historical data, motivating further exploration of sequential architectures designed to handle temporal data processing.

\subsection{Sequential Models Performance}

In this section, we assess the performance of sequential models, including LSTM, GRU, and TCN, in comparison to MLP, highlighting their effectiveness in capturing temporal dependencies and reducing compounding errors. Figure~\ref{fig:analysis_recursion} illustrates the mean and variance of composed predictions to evaluate the long-term predictive capability of these models on various unseen test trajectories. Notably, the mean and variance of MLP predictions are observed to be higher than those of the sequential models, indicating the ability of the latter to leverage temporal dependencies for more accurate predictions. This discrepancy arises primarily due to the inherent limitations of MLP architectures in capturing temporal dependencies effectively. As a result, MLPs struggle to leverage temporal context from historical information, leading to less accurate predictions over longer time horizons. Additionally, another factor contributing to the higher mean and variance in MLP predictions is the loss of causality in time-dependent signals. Unlike sequential models, which inherently preserve the temporal order of data through recurrent connections or $1$-D convolutional operations, MLPs process input data in a feedforward manner, disregarding the sequential nature of the information. This lack of causality can lead to discrepancies in predictions, especially over longer time horizons, where the relationships between data points play a critical role in accurate forecasting.
\begin{table*}[t]
    \setlength{\tabcolsep}{16.5pt}
    \centering
    \caption{Performance of predictor models with varied input state configurations. Given the critical role of control action in influencing system dynamics, it is included in all input configurations tested. Incorporating the full state information as input, consisting of linear velocity, angular velocity, attitude, and control action, results in the best performance. \\ *The attitude predictor is evaluated using the quaternion error.}
    \vspace{-0.25em}
    \begin{tabular}{ccccccccc}
    \toprule\toprule 
    \multicolumn{3}{c}{\rotatebox[origin=c]{0}{$\text{INPUT}$}} & \multicolumn{6}{c}{\rotatebox[origin=c]{0}{$\text{PREDICTOR}$}} \\
    \cmidrule(lr){1-3}\cmidrule(lr){4-9}
    \rotatebox[origin=c]{0}{$\mathbf{v}$} & \rotatebox[origin=c]{0}{$\boldsymbol{\omega}$} & \rotatebox[origin=c]{0}{$\mathbf{q}$} & 
    \multicolumn{2}{c}{\rotatebox[origin=c]{0}{$\text{$\mathbf{v}$}$}} & \multicolumn{2}{c}{\rotatebox[origin=c]{0}{$\text{$\boldsymbol{\omega}$}$}} & \multicolumn{2}{c}{\rotatebox[origin=c]{0}{$\text{$\mathbf{q}*$}$}}\\
    \cmidrule(lr){4-5}\cmidrule(lr){6-7}\cmidrule(lr){8-9}
    & & & PI-TCN & NeuroBEM & PI-TCN & NeuroBEM & PI-TCN & NeuroBEM \\
    \midrule
    \cmark & \xmark & \xmark & $0.102$ & $0.422$ & -- & -- & -- & -- \\
    \cmark & \cmark & \xmark & $0.072$ & $0.324$ & $0.162$ & $0.414$ & -- & -- \\
    \cmark & \xmark & \cmark & $0.014$ & $0.108$ & -- & -- & $0.071$ & $0.054$ \\
    \xmark & \cmark & \xmark & -- & -- & $0.194$ & $0.554$ & -- & -- \\
    \xmark & \cmark & \cmark & -- & -- & $0.121$ & $0.212$ & $0.061$ & $0.016$ \\
    \xmark & \xmark & \cmark & -- & -- & -- & -- & $0.086$ & $0.116$ \\
    \cmark & \cmark & \cmark & $\textbf{0.009}$ & $\textbf{0.072}$ & $\textbf{0.008}$ & $\textbf{0.091}$ & $\textbf{0.021}$ & $\textbf{0.006}$ \\
    \bottomrule\bottomrule
    \end{tabular}
    \label{tab:input_ablation}
    \vspace{-15pt}
\end{table*}

Furthermore, we observe that TCN outperforms all baseline models across various experiments, exhibiting superior performance in terms of predictive accuracy and stability. Moreover, our experiments reveal that TCN achieves a $21\times$ reduction in velocity error and a significant $23\times$ reduction in attitude error compared to MLP with no history on the PI-TCN dataset. Similarly, on the NeuroBEM dataset, TCN demonstrates a remarkable $31\times$ reduction in velocity error and an impressive $56\times$ reduction in attitude error compared to MLP with no history. These findings further underscore the superiority of sequential models, particularly TCN, in dynamics learning tasks, emphasizing their ability to capture complex temporal patterns and improve predictive accuracy.

\begin{figure}[!t]
  \begin{center}
    \includegraphics[width=\linewidth, trim=10 20 15 20, clip]{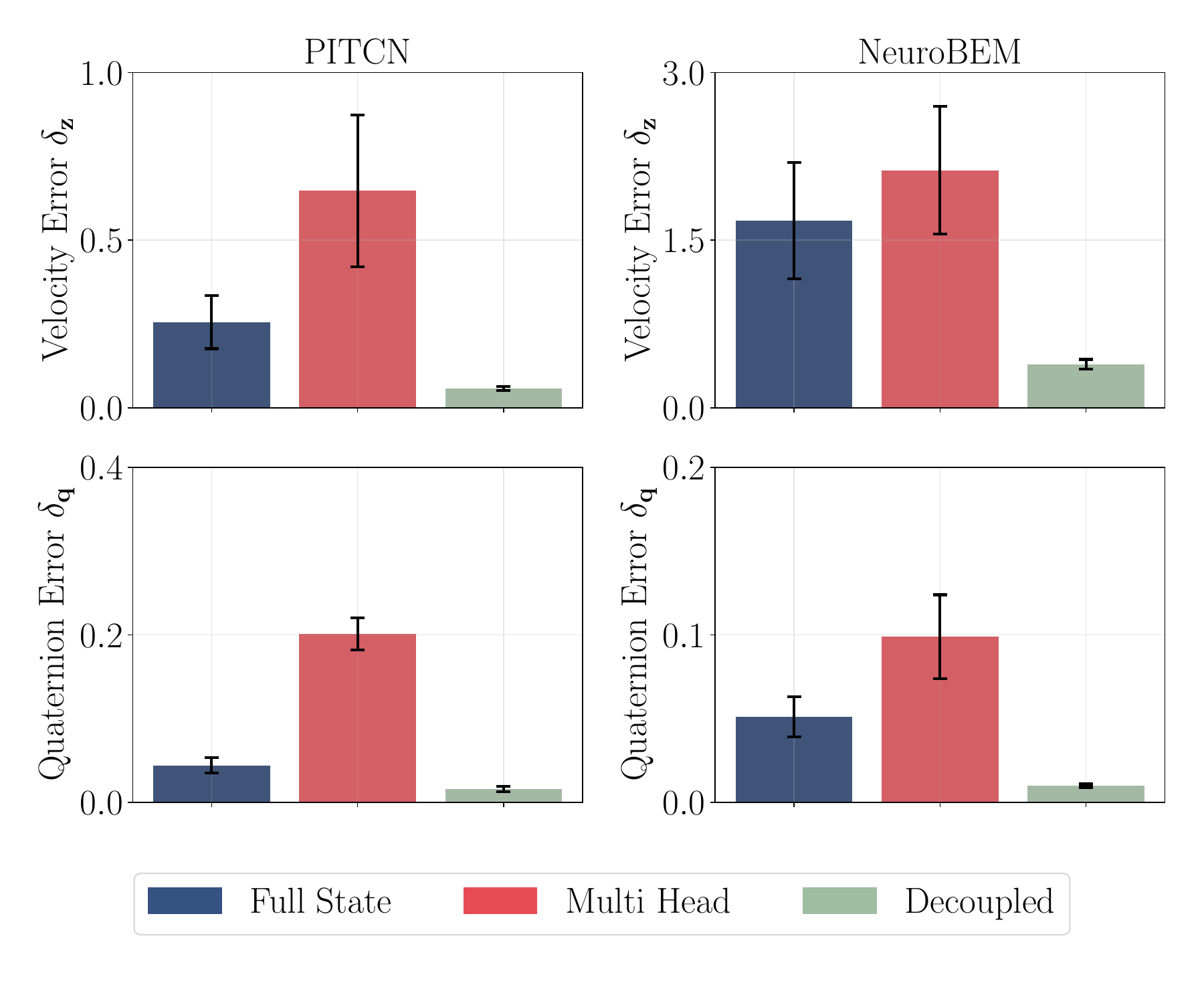}
    \caption{Mean and standard deviation of the best-performing predictors across $10$ random trials, evaluating velocity and attitude error. A, B, and C correspond to Full-State, Multi-Head, and Decoupled, respectively.}
    \label{predictor_ablation}
  \end{center}
  \vspace{-20pt}
\end{figure}

\subsection{Comparison of Predictor Types}
This section analyzes the predictive performance and stability of various predictor types across trajectories characterized by diverse levels of aggressiveness. These trajectories exhibit average velocities ranging from $1.12$ $\text{m}\text{s}^{-1}$ to $3.27$ $\text{m}\text{s}^{-1}$ for the PI-TCN dataset and $1.67$ $\text{m}\text{s}^{-1}$ to $15.02$ $\text{m}\text{s}^{-1}$ for the NeuroBEM dataset. Both velocity and attitude errors are assessed for each predictor type. Results indicate a consistent superiority of the decoupled dynamics predictor over both the full-state and multi-head predictors across all aggression levels, underlining its adeptness in handling challenging dynamic scenarios (refer Table~\ref{tab:agressive_flight_analysis}). Furthermore, an in-depth analysis of the mean and standard deviation of the best-performing predictors across $10$ different random seeds underscores the decoupled predictors' enhanced stability and robustness compared to their counterparts (refer Figure~\ref{predictor_ablation}). Notably, the instability and inaccuracy arises due to the complexity of the underlying problem. 

Additionally, we conduct an experiment to match the total number of parameters of the full-state and multi-head predictors to the decoupled predictor by doubling the parameters in both the encoder and decoder. This adjustment is necessary because the decoupled framework uses two independent neural networks to predict distinct modules, whereas the other two predictors rely on a single network. By doubling the parameters in the full-state and multi-head predictors, we aim to equalize model capacities. However, despite the increased model complexity, performance does not improve compared to the decoupled counterpart. This underscores the complex nature of the problem, where simply increasing model capacity does not enhance predictive accuracy. The decoupled approach resolves this by breaking down system dynamics into manageable subproblems, promoting modularity, simplifying learning, and enabling independent optimization for velocity and attitude components.

\subsection{Ablation of State Input Representation}
In this experiment, we aim to assess the impact of different input combinations on the performance of our framework for learning dynamics. We consider as inputs the history of velocity, angular velocity, attitude, and control action. Given the crucial role of control action in influencing system dynamics, it is included in all input combinations. We decompose the framework into three separate predictors: linear velocity, angular velocity, and attitude predictors to better understand the effects of different inputs on each solution.

Our observations reveal notable trends across the predictors (Table~\ref{tab:input_ablation}). Firstly, for the linear velocity predictor, we observe that incorporating the full state information as input, comprising the history of linear velocity, angular velocity, attitude, and control action, result in the best performance. Furthermore, there is a gradual increase in performance as we augment the input with additional state quantities, indicating the importance of considering the complete state information for accurate prediction. Similar trends are observed for the angular velocity and attitude predictors, with optimal performance achieved when full state observations are included as input. This ablation study highlights the significance of full state information in capturing the intricate dynamics of the system and achieving superior predictive accuracy.


\section{Discussion}
The proposed approach offers several new valuable insights and guidelines for designing data-driven solutions for quadrotor learning dynamics. First, leveraging historical information and employing multi-step loss formulation significantly enhances the model's ability to predict longer horizons. Second, the modular approach decouples system dynamics into manageable subproblems, facilitating independent optimization of each module. Finally, a TCN architecture integrated with the two aforementioned designs, demonstrates superior performance across diverse levels of aggressiveness in real-world scenarios. The top-performing model produces feasible and stable open-loop predictions for up to $60$ steps, showcasing its robustness and efficacy.

\vspace{-5pt}
\section{Conclusions} \label{sec:conclusion}
In summary, our work tackles the critical challenge of achieving accurate modeling of system dynamics to enable effective control of quadrotors. Despite the promises of existing data-driven approaches, their limitations in addressing compounding errors over long prediction horizons highlight the necessity for comprehensive strategies. Through meticulous evaluation of design choices and exploration of sequential modeling techniques, we demonstrate strategies in minimizing these errors. Our novel decoupled dynamics learning framework stands out for its ability to simplify the learning process while enhancing modularity, hence improving long-term forecasts. Extensive experiments on real-world datasets validate the efficacy and precision of our approach. Future work includes integrating the proposed framework with a controller to analyze flight performance under challenging operating conditions. 

\bibliographystyle{IEEEtran}
\bibliography{references}

\end{document}